\documentclass{article}

\usepackage{microtype}
\usepackage{graphicx}
\usepackage{subfigure}
\usepackage{booktabs} %
\usepackage{array}

\usepackage{amsmath}
\usepackage{amssymb}
\usepackage{amsfonts}

\usepackage[hyphens]{url} %
\usepackage{hyperref}

\usepackage{xcolor} %
\usepackage{pgfplots}
\usepackage{pgfplotstable}
\pgfplotsset{compat=1.16}

\usepackage[preprint]{mlsys2025}

\mlsystitlerunning{MXFP8 and MXFP6 Inference for Recommendation Models}

\begin{document}

\twocolumn[
\mlsystitle{Advancing Quantization for R\&R Models with Microscaling Formats}

\mlsyssetsymbol{equal}{*}

\begin{mlsysauthorlist}
\mlsysauthor{Yavuz Yetim}{meta}
\mlsysauthor{Bor-Yiing Su}{meta}
\mlsysauthor{Zhe Wang}{amd}
\mlsysauthor{Arthur Liu}{amd}
\mlsysauthor{Pranav Sharma}{meta}
\mlsysauthor{Ying Wang}{meta}
\mlsysauthor{Randy Shuai}{meta}
\mlsysauthor{Luke Pfister}{meta}
\mlsysauthor{Ruichao Xiao}{meta}
\mlsysauthor{Bill Zhao}{meta}
\mlsysauthor{Zihao Liu}{meta}
\mlsysauthor{Liyuan Li}{meta}
\mlsysauthor{Jiangyuan Li}{meta}
\mlsysauthor{Gaoxiang Liu}{meta}
\mlsysauthor{Nathan Yan}{meta}
\mlsysauthor{Geonhwa Jeong}{meta}
\mlsysauthor{Jatin Chhugani}{meta}
\mlsysauthor{Elias Ellison}{meta}
\mlsysauthor{Uladzimir Pashkevich}{meta}
\end{mlsysauthorlist}

\mlsysaffiliation{meta}{Meta Platforms, Inc., Menlo Park, USA}
\mlsysaffiliation{amd}{Advanced Micro Devices, Inc., Santa Clara, USA}

\mlsyscorrespondingauthor{Yavuz Yetim}{yyetim@meta.com}
\mlsyscorrespondingauthor{Bor-Yiing Su}{boryiingsu@meta.com}

\mlsyskeywords{Machine Learning, MLSys}

\vskip 0.3in

\begin{abstract}
Recommendation and Ranking (R\&R) models, particularly in digital advertising, require massive inference throughput
but exhibit extreme sensitivity to precision loss. Unlike Large Language Models (LLMs) that frequently tolerate quantization-induced
accuracy drops of 1\% or more, production-grade R\&R models enforce a stringent launch bar:
model quality must remain within a 0.01\% Normalized Entropy (NE) margin relative to the baseline, which commonly uses FP16 or FP32.
This tight constraint has historically hindered the adoption of aggressive quantization in this domain.
In this paper, we present an empirical evaluation of both MXFP8 and the 6-bit MXFP6 format for industrial-scale R\&R inference on the AMD MI350X architecture.

\textbf{To the best of our knowledge, this is the first paper, across R\&R and LLM domains, that reports significant speedups from native MXFP6 GEMMs, which quantize both weights and activations and compute directly on 6-bit elements, without sacrificing model quality}. We demonstrate that we can selectively apply this 6-bit format to maximize overall QPS gain without breaching the strict NE parity threshold. We further show that MXFP8 is a viable alternative to row-wise FP8: the two formats have nearly the same accuracy on our models, so the faster kernel can be chosen for each GEMM shape.
On two production ads ranking models, our implementation shows measured GEMM kernel speedups over FP16 of $1.33$--$1.55\times$ for MXFP8 and $1.73$--$1.99\times$ for MXFP6 across all supported linear layers, and up to $2.90\times$ for MXFP6 on individual layers,
alongside end-to-end (E2E) inference throughput gains from selective MXFP6 of $1.5$--$5.3\%$ for Model A and $7.6$--$14.3\%$ for Model B over their FP8 configurations.
\end{abstract}

]

\printAffiliationsAndNotice{} %

\section{Introduction}
\label{sec:intro}

The rapid evolution of artificial intelligence has been heavily driven by the scaling of model sizes and compute requirements. While Large Language Models (LLMs) have dominated the discourse on compute scaling, Recommendation and Ranking (R\&R) models are scaling at a pace that was unthinkable only a few years ago~\cite{luo2026loka}. In Meta's ads ranking stack, for instance, the inference work per example of the most complex launched late-stage ranking model reached approximately $22.2\times$ its first-half 2024 level in the first half of 2026, with an estimated target of $33.3\times$ that same starting point in the second half (Figure~\ref{fig:rr_scaling})~\cite{meta2026adaptiveranking}.

Unlike LLM serving, however, ads ranking must return results within a strict sub-second latency budget and with a computational cost that the model quality gains can justify. Therefore, every increase in model complexity has to be paid for with higher hardware efficiency. However, scaling hardware efficiency is increasingly bottlenecked by power consumption limits. To deliver higher throughput with contained power consumption, modern GPU architectures offer throughput gains primarily through lower-precision formats. For example, accelerators such as NVIDIA's Blackwell (GB300)~\cite{blackwell_spec} and AMD's MI350X~\cite{amd_mi350x_brochure} introduce native support for sub-8-bit computing. By supporting the Open Compute Project (OCP) Microscaling Formats (MX)~\cite{ocp2023mx}, the AMD MI350X architecture offers a theoretical peak compute scaling ratio of 1:2:4 as precision scales down from FP16 to MXFP8 and MXFP6 (Table~\ref{table:hw_throughput}). To sustain the compute scaling of R\&R models without violating strict latency and computational cost budgets, we aim to take advantage of this efficiency gain opportunity and adopt these lower-precision formats.

\begin{table}[ht!]
  \centering
  \caption{Dense peak tensor throughput of each precision, relative to FP16~\cite{blackwell_spec,amd_mi350x_brochure}. MI350X has twice the MXFP8 peak throughput for MXFP6.}
  \label{table:hw_throughput}
  \begin{tabular}{@{}lcc@{}}
    \toprule
    Precision     & NVIDIA GB300 & AMD MI350X \\
    \midrule
    FP16          & $1\times$    & $1\times$  \\
    FP8 / MXFP8   & $2\times$    & $2\times$  \\
    MXFP6         & $2\times$    & $4\times$  \\
    MXFP4         & $6\times$    & $4\times$  \\
    NVFP4         & $6\times$    & --         \\
    \bottomrule
  \end{tabular}
\end{table}

Despite these theoretical hardware benefits and the widespread success of low-precision inference in LLMs—where quantization studies routinely describe accuracy drops of 1--3\% as modest~\cite{kurtic2025bf16} and industry benchmarks accept results within 99\% of the FP16 reference~\cite{mlcommons_inference_rules}—R\&R models have been notoriously resistant to aggressive quantization. In the highly competitive domain of digital advertising, R\&R models are extremely sensitive to any degradation in prediction quality. The established launch criterion for these models is exceptionally stringent: any proposed optimization must maintain a Normalized Entropy (NE) difference of no more than $0.01\%$ compared to the production baseline. This near-zero tolerance for quality loss has historically limited sub-16-bit quantization in production R\&R systems to a few selected layers, leaving a massive gap between theoretical hardware capabilities and actual deployed efficiency~\cite{luo2026loka}.

In this paper, we bridge this gap by demonstrating that microscaling formats can unlock high-throughput inference for R\&R models: selective MXFP6 stays within the strict $0.01\%$ NE launch bar, and MXFP8 has nearly the same accuracy as row-wise FP8. To the best of our knowledge, this is the first work to deploy MXFP6 in production-grade recommendation models within this launch bar. Our key contributions are summarized as follows:

\begin{enumerate}
    \item \textbf{Tensor-Level Quantization Analysis:} We measure the quantization signal-to-noise ratio (QSNR) of every supported linear layer of two production models, using real production inputs and the actual low-precision kernels. MXFP8 has nearly the same QSNR as row-wise FP8, and the median per-call-site gap between MXFP6 and MXFP8 is only about 1~dB.
    
    \item \textbf{Kernel and End-to-End Optimizations:} We develop MXFP8 and MXFP6 GEMM kernels for the AMD MI350X and fuse activation quantization into the producer kernels. To the best of our knowledge, this is the first work to report native MXFP6 GEMM speedups: $2.21$--$2.31\times$ over FP16 on the 20 most compute-intensive linear layers of each model and up to $2.90\times$ on individual layers ($1.67$--$1.72\times$ for MXFP8). In full-model benchmarks at batches 1024 and 2048, selective MXFP6 improves end-to-end throughput over the FP8 configurations by $5.3\%$ and $1.5\%$ for Model A and $14.3\%$ and $7.6\%$ for Model B, respectively.
    
    \item \textbf{A New Quantization Paradigm for R\&R Models:} To the best of our knowledge, this is the first work to push R\&R quantization below 8 bits within the strict $0.01\%$ NE launch bar. MXFP6 has the lowest aggregate GEMM time, and MXFP8 is a viable alternative to row-wise FP8, chosen per GEMM shape by kernel speed.
\end{enumerate}

\section{Background}
\label{sec:background}

\subsection{R\&R Model Architecture}
\label{sec:background:arch}

Industrial R\&R models follow the general structure of DLRM~\cite{naumov2019dlrm}.
Categorical (sparse) features, such as user and ad identifiers, are mapped to dense vectors through lookups into large embedding tables,
while continuous (dense) features are projected into the same embedding space by an MLP.
An interaction architecture then models the relationships among these embeddings,
and an MLP produces the final predictions, often for many tasks at once.
Embedding tables hold the vast majority of the parameters~\cite{meta2026gemtraining},
but embedding lookups are memory-bandwidth bound and contribute few FLOPs,
so the dense compute is concentrated in the interaction and MLP layers and is dominated by GEMMs.
These GEMMs are the target of the quantization techniques we study in this paper.

Meta has published a line of increasingly deep interaction architectures for ads ranking.
DHEN~\cite{zhang2022dhen} stacks layers that each ensemble heterogeneous interaction modules, such as the Deep \& Cross Network (DCN)~\cite{wang2021dcnv2},
and Wukong~\cite{zhang2024wukong} streamlines this design into a stack of identical layers that exhibits a scaling law across two orders of magnitude of model complexity.
R\&R models also consume sequences of past user interactions;
InterFormer~\cite{zeng2024interformer} interleaves transformer-style sequence learning with the interaction of the non-sequence features.
The ads foundation model GEM~\cite{meta2025gem}, trained at LLM scale~\cite{meta2026gemtraining}, transfers its learnings to the models that serve ads,
which run under inference latency and capacity budgets.

\subsection{Scaling R\&R Inference}
\label{sec:background:scaling}

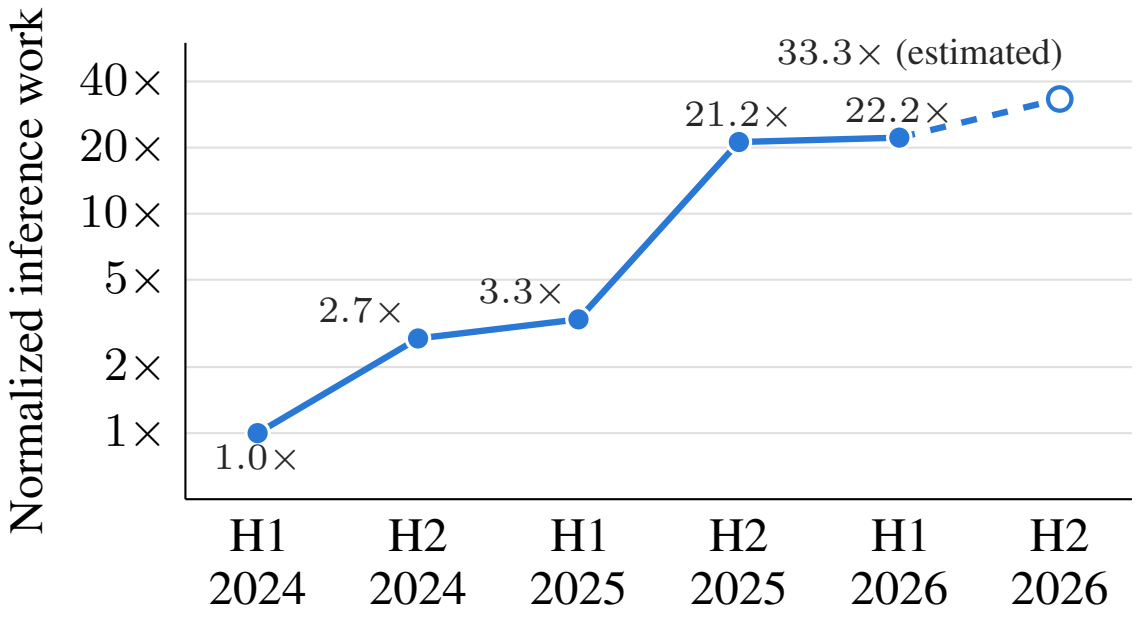
\begin{figure}[t]
\centering
\definecolor{series1}{HTML}{2A78D6}
\begin{tikzpicture}
\begin{axis}[
    width=0.95\columnwidth, height=4.6cm,
    ymode=log, log basis y=10,
    ymin=0.5, ymax=60,
    ytick={1,2,5,10,20,40},
    yticklabels={$1\times$,$2\times$,$5\times$,$10\times$,$20\times$,$40\times$},
    ylabel={Normalized inference work},
    ylabel style={font=\footnotesize},
    xtick={0,...,5},
    xticklabels={H1 2024,H2 2024,H1 2025,H2 2025,H1 2026,H2 2026},
    enlarge x limits=0.09,
    axis x line*=bottom, axis y line*=left,
    ymajorgrids, grid style={line width=0.4pt, draw=black!12},
    tick label style={font=\footnotesize},
    xticklabel style={text width=2em, align=center},
    tick style={draw=none},
    nodes near coords={\pgfmathprintnumber[fixed, fixed zerofill, precision=1]{\pgfplotspointmeta}$\times$},
    nodes near coords style={font=\scriptsize, text=black!85, anchor=south, inner sep=2pt},
    point meta=rawy,
]
\addplot[draw=series1, line width=1.2pt, mark=*, mark size=2.2pt,
    mark options={fill=series1, draw=white, line width=0.6pt},
    nodes near coords={}]
    coordinates {(0,1.0) (1,2.7) (2,3.3) (3,21.2) (4,22.2)};
\addplot[draw=none, mark=none, nodes near coords style={anchor=north}]
    coordinates {(0,1.0)};
\addplot[draw=none, mark=none, nodes near coords style={anchor=south east}]
    coordinates {(1,2.7) (2,3.3)};
\addplot[draw=none, mark=none, nodes near coords style={anchor=south}]
    coordinates {(3,21.2) (4,22.2)};
\addplot[draw=series1, dashed, line width=1.2pt, mark=none, nodes near coords={}]
    coordinates {(4,22.2) (5,33.3)};
\addplot[draw=none, mark=*, mark size=2.2pt, mark options={fill=white, draw=series1, line width=0.8pt, solid},
    nodes near coords={\pgfmathprintnumber[fixed, fixed zerofill, precision=1]{\pgfplotspointmeta}$\times$ (estimated)},
    nodes near coords style={anchor=south east, xshift=3pt, yshift=3pt}]
    coordinates {(5,33.3)};
\end{axis}
\end{tikzpicture}
\caption{Inference work per example of the most complex launched late-stage ads ranking model at Meta, normalized to H1'24 ($1\times$), on a logarithmic scale.
The sharp increase in H2'25 coincides with the first launch of the Adaptive Ranking Model~\cite{meta2026adaptiveranking}; H2'26 is an estimated target.}
\label{fig:rr_scaling}
\end{figure}

We report inference complexity as work per example (one user--ad pair), normalized to the first half of 2024 (Figure~\ref{fig:rr_scaling}).
Before adaptive ranking, every model had to serve all traffic within the same latency and capacity budget,
so a larger model raised the computational cost of every request, and this work grew to only approximately $3.3\times$ the starting level by the first half of 2025.
The Meta Adaptive Ranking Model~\cite{meta2026adaptiveranking} changed this trajectory.
It routes requests to models of different complexity based on the person's context and intent,
so that the most complex models are spent on the requests where they deliver the most value,
and it co-designs these models with the serving system to keep their computational cost sub-linear in complexity,
for example by computing user-side signals once per request instead of once per user--ad pair.
Its first iterations launched in the second half of 2025 and brought the frontier to approximately $21.2\times$ the starting level;
complexity reached $22.2\times$ in the first half of 2026, with an estimated target of $33.3\times$ in the second half.
Because the added compute is concentrated in the dense interaction and MLP layers,
an increasing share of inference time goes to GEMMs, which makes them the natural target for quantization.

\subsection{Common Quantization Formats}

\begin{table*}[t]
\caption{Low-precision formats discussed in this paper. All formats have one sign bit. Bits per element include the amortized storage of the scales; FP16 is the baseline. MXFP8 and MXFP6 are listed with their high-mantissa variants (E4M3 and E2M3, respectively).}
\label{tab:formats}
\centering
\small
\setlength{\tabcolsep}{3pt}
\begin{tabular}{@{}lccccccl>{\raggedright\arraybackslash}p{2.0cm}c@{}}
\toprule
Format & Exp. & Man. & $F_{\max}$ & $F_{\min}$ & Dyn. Range & Scale & Granularity & Bits \\
\midrule
FP16              & 5   & 10  & $65{,}504$ & $2^{-24}$ & ${\approx}1.10 \times 10^{12}$ & --          & --                                   & 16 \\
FP8 (tensor)      & 4   & 3   & $448$   & $2^{-9}$  & $229{,}376$                      & FP32        & Whole tensor                         & 8 \\
FP8 (row)         & 4   & 3   & $448$   & $2^{-9}$  & $229{,}376$                      & FP32        & Row (act.), channel (wt.)            & ${\approx}8$ \\
MXFP8             & 4   & 3   & $448$   & $2^{-9}$  & $229{,}376$                      & E8M0        & Block of 32                          & 8.25 \\
MXFP6             & 2   & 3   & $7.5$   & $2^{-3}$  & $60$                           & E8M0        & Block of 32                          & 6.25 \\
NVFP4             & 2   & 1   & $6$     & $2^{-1}$  & $12$                           & E4M3 + FP32 & Block of 16 + tensor                 & 4.5 \\
MXFP4             & 2   & 1   & $6$     & $2^{-1}$  & $12$                           & E8M0        & Block of 32                          & 4.25 \\
\bottomrule
\end{tabular}
\end{table*}

Floating-point (FP) numbers are composed of one sign bit, $E$ bits for exponents, and $M$ bits for mantissas. The definitions of the common low-precision types we consider are summarized in Table~\ref{tab:formats}. As all common numerical types have a sign bit, it is typically omitted when referencing low-precision types. For example, the FP16 format is E5M10, with 5 bits of exponent and 10 bits of mantissa, and the MXFP6 format is E2M3, with 2 bits of exponent and 3 bits of mantissa. The exponent bits set the dynamic range of an element, while the mantissa bits determine its precision (the number of bins that differentiate values between two consecutive exponent boundaries).

To preserve more information, these formats pair elements with metadata, most commonly a scaling factor, that maps groups of values into the representable range. The dynamic range of a format is defined as $F_{\max} / F_{\min}$, where $F_{\max}$ is the absolute maximum representable value and $F_{\min}$ is the absolute minimum positive subnormal value. Values beyond these bounds must be saturated to $F_{\max}$ or rounded to 0, and the scaling factor reduces such overflow and underflow by fitting most values into the dynamic range. Formats differ mainly in the size of the group sharing this scale. FP8~\cite{micikevicius2022fp8} with tensor-wise scaling uses one FP32 scale per tensor, meaning a single outlier reduces the effective precision of every other value in the tensor. Row-wise scaling~\cite{mor2025} keeps one scale per row of the activation matrix and one per output channel of the weight matrix, confining the effect of an outlier to its row or channel.

The OCP~\cite{ocp2023mx, rouhani2023mx} microscaling (MX) formats go further and share one scaling factor across each block of 32 consecutive elements along the reduction dimension of the GEMM, for 8-bit (MXFP8), 6-bit (MXFP6), and 4-bit (MXFP4) elements. The MX scale uses an 8-bit power-of-two (E8M0) format with no sign bit, so applying it simply shifts exponents. The E8M0 scale has sufficient dynamic range for any practical outliers, but it lacks the precision to preserve the per-block absolute maximum $\alpha_{\max}$, which introduces more quantization error for 4-bit elements. NVIDIA's NVFP4 format~\cite{nvidia2025nvfp4} therefore reduces the block size to 16 elements and stores its scales in E4M3, with an additional per-tensor FP32 scale to compensate for the narrower range of E4M3~\cite{nvfp4_training}.

\subsection{Quantization in Production R\&R Models}
\label{sec:background:quant_in_rr}

Before this work, low-precision inference in Meta's ads ranking models was limited to FP8 on a few selected linear layers.
Using LoKA~\cite{luo2026loka}, which benchmarks low-precision GEMM kernels on the shapes and input distributions of production ads models,
engineers identified the linear layers that tolerate FP8 and switched only those to FP8 GEMMs with tensor-wise or row-wise scaling;
all other layers stayed in FP16, the baseline format for R\&R inference.
The Adaptive Ranking Model uses the same selective approach, deploying FP8 post-training quantization only in layers with high tolerance to precision loss~\cite{meta2026adaptiveranking}.
Because only some linears are quantized, a large share of GEMM compute still runs in FP16,
which leaves much of the low-precision throughput of current GPUs unused.

\section{MXFP8 as an Alternative 8-bit Format}
\label{sec:mxfp8}

\subsection{MXFP8 Representation Fidelity}
\label{subsec:mxfp8_fidelity}

The OCP MXFP8 specification~\cite{ocp2023mx} defines two distinct 8-bit floating-point formats: E5M2 (5 exponent bits, 2 mantissa bits) and E4M3 (4 exponent bits, 3 mantissa bits). 
The E5M2 format offers a wide dynamic range, with $F_{\max} = 57,344$ and $F_{\min} = 2^{-16}$, yielding a dynamic range of approximately $3.76 \times 10^9$. In contrast, E4M3 trades exponent range for mantissa precision, providing $F_{\max} = 448$ and $F_{\min} = 2^{-9}$, which results in a dynamic range of $229{,}376$. 

Under the microscaling (MX) paradigm, every block of 32 consecutive elements shares an 8-bit E8M0 scale factor. Because this local scale dynamically aligns the exponent range to the peak magnitude within each 32-element vector, the intra-block dynamic range rarely exceeds the capacity of $229{,}376$. Consequently, the extended exponent range of E5M2 becomes redundant at a block size of 32, making mantissa precision the dominant factor in minimizing quantization error.

We measure quantization error with the quantization signal-to-noise ratio (QSNR)~\cite{rouhani2023bdr, mxfp4_oas}. For a reference tensor $y_{\mathrm{ref}}$ and its low-precision counterpart $y$,
\begin{equation}
\mathrm{QSNR} = 10 \log_{10} \frac{\|y_{\mathrm{ref}}\|^2}{\|y - y_{\mathrm{ref}}\|^2},
\end{equation}
where the norms are taken over all elements, so QSNR is the ratio of signal power to quantization noise power in decibels.
The single additional mantissa bit in E4M3 ($M=3$) over E5M2 ($M=2$) yields a significant theoretical improvement in QSNR. Under standard quantization noise theory~\cite{oppenheim2009discrete, gray1998quantization}, each additional mantissa bit halves the width of the quantization bins and increases QSNR by approximately $6.02\text{ dB}$ (Appendix~\ref{sec:appendix:qsnr_bit}).
Because recommendation models are highly sensitive to small perturbations, this $\approx 6\text{ dB}$ QSNR gain renders E4M3 substantially superior to E5M2 for preserving model quality in MXFP8 execution. This is why we only consider E4M3 format in our studies.

We reduce MXFP8 quantization error with the Overflow-Aware Scaling (OAS) algorithm of~\citet{mxfp4_oas}, adapted for E4M3. OAS chooses the E8M0 scale to accept a small saturation of the block maximum in exchange for doubling the representable dynamic range for the lower-magnitude elements of the affected blocks. Because E4M3 reserves the all-ones encoding, which would otherwise be its largest value (480), for NaN, we use Overflow-Aware Scaling Skip (OAS-S), which skips this encoding during scale rounding and bounds the relative error from saturating the block maximum to $3.4\%$. Appendix~\ref{sec:appendix:oas} describes OAS-S in detail.

\subsection{MXFP8 Kernel Performance}
\label{subsec:mxfp8_perf}

\textbf{GEMM kernels.}
We designed and implemented our MXFP8 and MXFP6 GEMM kernels for MI350X in FlyDSL~\cite{amd_flydsl}.
Each kernel offers several configurations, each a tiling and pipeline design that we implemented,
and during development the AI agent AlphaKernel~\cite{meta_kernelevolve_2026} tuned the parameters of these configurations for the GEMM shapes of our production models.
Our kernels reach and exceed the performance of vendor kernels on our production shapes.
The row-wise FP8 baseline in our measurements calls \texttt{torch.\_scaled\_mm}, which runs AMD's hipBLASLt FP8 GEMM.
Summed over the 20 most compute-intensive layers, our MXFP8 kernel is faster than FP8 at batch 1024 and slower at batch 2048.
Our MXFP6 kernel, described in Section~\ref{subsec:mxfp6_perf}, is fastest on every shape in that subset at both batch sizes and has the lowest summed GEMM time across all supported linear layers (Section~\ref{sec:eval:all}).
For the large shapes that dominate the GEMM time, the speedups thus order as FP16 $<$ FP8 $\approx$ MXFP8 $<$ MXFP6, as the peak throughputs of the formats predict.

\textbf{Activation preparation.}
A faster GEMM only speeds up inference if its inputs can be prepared efficiently.
Before each MX GEMM, the activation must be quantized:
for each block of 32 elements, we compute the maximum absolute value, derive the E8M0 scale,
scale and clamp the elements, convert them to the element format,
pack them into bytes, and write the block scales in the layout the kernel expects.
Run as separate kernels, these steps read and write the activation several times and outweigh the GEMM gains.
In an earlier sweep with standalone quantization and untuned MXFP6 GEMMs,
the summed time of the measured linears at batch 1024 is $1.7\times$ (MXFP6) and $4.6\times$ (MXFP8) the FP16 time for Model A,
and $1.6\times$ (MXFP6) and $5.2\times$ (MXFP8) for Model B.
These measurements include activation preparation, unlike the tuned GEMM-only speedups in Table~\ref{tab:eval_all}.

Our production inference stack generates the fused kernels around linear layers with PyTorch Inductor,
but Inductor did not support storing block scales and sub-byte element types efficiently.
We extended Inductor so that the quantization operations, including abs, min, max, mul, clamp, and byte packing,
are appended to the chain of producer operations of each activation and fused with them.
The activation is thus quantized while it is produced and written to memory only once, already packed and with its block scales.
We tested the generated kernels against our own hand-tuned activation fusions to identify and remove performance bottlenecks.
As a result, the activation preparation for MXFP8 and MXFP6 is at least as efficient as the row-wise FP8 preparation used in production today.

\section{MXFP6 as the New Aggressive Candidate}
\label{sec:mxfp6}

\subsection{MXFP6 Representation Fidelity}
\label{subsec:mxfp6_fidelity}

Similar to the MXFP8 format discussed in Section~\ref{subsec:mxfp8_fidelity}, the OCP MXFP6 specification defines two distinct 6-bit configurations: E3M2 (3 exponent bits, 2 mantissa bits) and E2M3 (2 exponent bits, 3 mantissa bits). The E3M2 format offers a wide dynamic range of $448$ ($F_{\max} = 28$, $F_{\min} = 2^{-4}$), whereas the E2M3 format provides a dynamic range of $60$ ($F_{\max} = 7.5$, $F_{\min} = 2^{-3}$). Although E2M3 has a narrower dynamic range than E3M2, it has the same three mantissa bits as E4M3. Because the E8M0 scale is a power of two, scaling only shifts exponents and does not change how the mantissa of an element is rounded. Every element in the normal range of E2M3 is therefore quantized to the same value as in MXFP8 (E4M3). Only elements in the subnormal range of E2M3, more than about 4--8$\times$ smaller than the block maximum, lose precision relative to MXFP8. This makes E2M3 the superior choice for model quality. There is no NaN encoding in the E2M3 format, so we apply the regular OAS algorithm~\cite{mxfp4_oas} to the MXFP6 scaling factors; Appendix~\ref{sec:appendix:oas} gives the details.

\subsection{MXFP6 Kernel Performance}
\label{subsec:mxfp6_perf}

We target a heterogeneous set of production linears, illustrated by the shapes in Table~\ref{tab:eval_top20}.
They include MLPs with wide inputs, projections with wide outputs, and low-rank down- and up-projections with strongly different aspect ratios.
The row dimension also varies with batch size and whether a layer runs per request or per candidate ad.
These differences affect parallelism, operand reuse, and memory access, motivating shape-specific tiling, pipeline configurations, and split-K work distribution.

The main challenge of an efficient MXFP6 GEMM is its element size.
An MXFP6 element effectively occupies 0.75 bytes:
four elements share three bytes, and a block of 32 elements occupies 24 bytes, which is not a power of two.
GEMM kernels are normally built around power-of-two data sizes, and MXFP6 breaks this assumption.
The byte address of an element no longer follows from a shift of its index,
a 16-byte vector load holds a fractional number of elements,
and tile dimensions that are powers of two in elements are multiples of three in bytes.

Most of our effort therefore went into fitting multiple-of-three data sizes into each level of the storage hierarchy while keeping the matrix cores busy.
We designed how tiles are read from global memory into the local data share (LDS),
how the LDS prefetch buffers are laid out and sized,
how the packed operand fragments are assembled in vector general-purpose registers (VGPRs) for the matrix instructions,
and how the software pipeline overlaps the global loads, the LDS reads, and the matrix instructions.
The rest of this section describes the resulting kernel.

\textbf{Matrix instruction and tiling.}
The kernel is built around the scaled matrix instruction of MI350X,
which computes a $32 \times 32$ output tile over 64 values of $K$ for one 64-lane wave, with both operands in E2M3.
Each lane supplies a 24-byte fragment of each operand containing one 32-element MX quantization block, together with its E8M0 scale byte.
We process $K$ in K-tiles of 192 elements, which are 144 bytes per row,
and split each K-tile into three sub-tiles of 64 elements, each requiring one matrix instruction per output tile.
A workgroup of 256 threads, four waves with one per SIMD, computes an output region of 128 or 256 rows by 64 to 512 columns.
The waves form a $2 \times 2$ or $1 \times 4$ grid, and each wave computes up to $4 \times 4$ output tiles.
Each output tile requires 16 FP32 accumulators per lane; at the largest wave tile, these occupy all 256 accumulation registers per lane.
The large accumulator tiles leave room for few waves per SIMD, so the kernel hides memory latency by software pipelining rather than by occupancy.
Within a K sub-tile, a wave reads each A fragment once and reuses it across its output-tile columns, and reuses each B fragment across its output-tile rows.

\textbf{Operand staging.}
The two operands take different paths through the storage hierarchy.
The activations (A) go through LDS:
every row of a workgroup tile is needed by all waves in a wave row,
and a fragment needs 24 bytes from each of 32 rows, which would be an uncoalesced access pattern in global memory.
Each thread moves 16-byte chunks with direct-to-LDS loads that bypass the registers.
A 144-byte tile row thus takes nine chunks, and the tile is stored in LDS with a 144-byte row pitch,
from which a lane assembles its 24-byte fragment with one 16-byte and one 8-byte LDS read.
The weights (B) bypass LDS.
We preshuffle them offline so that each pair of a 32-column segment and a 64-element K sub-tile forms one contiguous 1,536-byte record in the lane order of the matrix instruction.
A B fragment is then one fully coalesced 16-byte and 8-byte load per lane, already in operand layout,
and the kernel keeps B in a ring of prefetched fragments in registers,
which saves LDS capacity, barriers, and an LDS round trip per fragment.
The scales also go directly to registers.
They are grouped by a wave's output-tile rows and columns, offline for the weights and by the activation preparation for the activations,
so that each lane loads its scales for a K-tile with one 12-byte load per operand, one byte per K sub-tile and output-tile row or column;
this grouping limits a wave to four output tiles per dimension.

\begin{algorithm}[t]
\caption{Main loop of the deep-pipelined MXFP6 GEMM for one workgroup. K-tile $t$ uses LDS slot $t \bmod 3$.}
\label{alg:mxfp6_deep}
\begin{algorithmic}[1]
\STATE zero the accumulators; prefetch the initial B ring and the scales of tile 0
\STATE DMA the A tiles 0 and 1 into LDS slots 0 and 1; wait; read the A fragments of sub-tile 0 of tile 0
\FOR{$t = 0, \dots, T-2$}
  \FOR{each sub-tile $s = 0, 1, 2$ of tile $t$}
    \IF{$s = 0$}
      \STATE load the scales of tile $t{+}1$
    \ENDIF
    \IF{$s = 2$}
      \STATE wait until at most $N_{\mathrm{wait}}$ loads are outstanding and all LDS reads are done; barrier
    \ENDIF
    \STATE read the A fragments of the next sub-tile from LDS (sub-tile $s{+}1$ of tile $t$, or sub-tile 0 of tile $t{+}1$ from slot $(t{+}1) \bmod 3$ if $s = 2$)
    \FOR{each N block of the wave}
      \STATE issue the matrix instructions for all output-tile rows with the next B fragment of the ring
      \STATE refill that ring entry with the B fragment $d$ positions ahead, in tile $t$ or $t{+}1$
      \STATE issue this position's share of the DMA of A tile $t{+}2$ into slot $(t{+}2) \bmod 3$
    \ENDFOR
  \ENDFOR
\ENDFOR
\STATE compute the last K-tile (peeled out of the loop); fence the accumulators; store the output tile
\end{algorithmic}
\end{algorithm}

\textbf{Deep pipelining.}
Our first MXFP6 kernel used two activation buffers in LDS and waited for all outstanding loads at every K-tile boundary.
It then issued new loads before resuming matrix instructions, leaving the matrix cores idle and making it slower than vendor FP8 GEMM on large-$K$ shapes.
The deep-pipelined kernel (Algorithm~\ref{alg:mxfp6_deep}) overlaps these loads with computation.
It uses three LDS buffers for the current, next, and following activation tiles.
While computing tile $t$, it distributes loads for tile $t{+}2$ throughout the matrix instructions.
Weight fragments are prefetched into a register ring, keeping 6--12 fragments ahead even across K-tile boundaries.

Each K-tile has one workgroup barrier, just before its final sub-tile.
A counted wait ensures that the next activation tile is ready without waiting for all newer loads to finish.
The kernel also waits for outstanding LDS reads and synchronizes the waves before reusing a buffer.
It then reads the next tile's first activation fragments while computing the current tile's final sub-tile.
Additional instruction waits protect the inline-assembly operands, and a final accumulator fence ensures that the output sums are ready before storing them.

The main loop avoids branches to keep accumulators in their dedicated registers.
Surplus and tail loads use valid addresses and reserved LDS padding instead of conditional skips.
We handle the final K-tile separately; otherwise, the compiler copies accumulators between register types on every iteration, increasing GEMM time by about 25\%.
Finally, the 144-byte row segment of a K-tile straddles 128-byte cache lines in a row-major activation matrix.
Storing the activations K-tile-major makes each activation tile contiguous in memory and improves the deep-pipelined kernel by about 6--8\%;
the activation preparation has to produce this layout.

\textbf{Work distribution.}
MI350X assigns consecutive workgroups to its eight XCDs in a round-robin order, and each XCD has a private L2 cache.
With the default order, neighboring tiles that share a column panel of B would stream it through all eight L2 caches.
Some configurations therefore remap the workgroup indices so that each XCD processes a contiguous run of tiles in column-major order, which share their B panels in its L2 cache.
When the output grid is too small to fill the 256 compute units, for example for a small $N$ with a large $K$, we split $K$:
the K-tiles are divided into segments, each segment writes an FP32 partial result,
and a separate kernel sums the partials, adds the bias, and converts the result to the output type.
$M$ is a runtime argument, so each kernel is compiled once per $K$, $N$, and configuration and reused for any batch size.
The hardware drops stores beyond the bounds of the output buffer, so the epilogue needs no per-element guards,
and rows beyond $M$ in the last tile simply re-read the last valid row.
For $K \le 576$, a single-wave kernel without LDS loads both operands directly into registers.
$K$ is zero-padded to a multiple of the 192-element K-tile, and the weights are zero-padded so that $N$ is a multiple of the output tile, with the output sliced back to $N$;
the kernel therefore supports any GEMM shape.

\textbf{Configurations.}
Each combination of tile shape, pipeline, prefetch distance, workgroup order, and schedule options is one configuration of the kernel.
As for MXFP8 (Section~\ref{subsec:mxfp8_perf}), AlphaKernel tuned the parameters of these configurations,
and for each production shape we use the fastest configuration together with a split-K factor (Section~\ref{sec:eval:method}).

With these kernels, MXFP6 has the lowest aggregate GEMM time of the three formats (Section~\ref{sec:eval:all}).
The activation preparation of Section~\ref{subsec:mxfp8_perf} applies to MXFP6 as well,
with byte packing that stores every four elements in three bytes.

\section{Evaluation}
\label{sec:evaluation}

\subsection{Experimental Methodology}
\label{sec:eval:method}

\textbf{Model and data.}
All experiments use two real production ads ranking models, which we call Model A and Model B.
The per-linear studies use real production data, not synthetic inputs or randomly initialized weights;
the end-to-end benchmark inputs are described in Appendix~\ref{sec:appendix:e2e}.
For each model, we captured the inputs of its dense network from production inference traffic.
For Model A, we captured 53 batches with 69 to 638 candidate ads each, for a total of about 17.9k user--ad examples;
its dense network contains 313 linear-layer call sites with 47 unique GEMM shapes.
For Model B, we captured 63 batches with 39 to 877 candidate ads each (about 19.9k examples);
its dense network contains 202 linear-layer call sites with 46 unique GEMM shapes.
We describe a GEMM by its reduction dimension $K$, its output dimension $N$, and its number of rows $M$
(the batch size, or the number of requests for request-level layers, times the rows per example).
We replay the captured inputs through every linear layer and compare its output in each format.
Two layers of Model B have a reduction dimension $K$ that is not a multiple of 32.
We zero-pad their $K$ to the next multiple of 32, and we time all layers with the padded $K$ in every format, including FP16.

\textbf{Formats.}
We compare four formats (Table~\ref{tab:formats}):
FP16, the production baseline;
row-wise FP8 (E4M3), the scheme used for selected layers in production today (Section~\ref{sec:background:quant_in_rr});
MXFP8 with E4M3 elements;
and MXFP6 with E2M3 elements.
Activations are quantized dynamically for every call.
The FP8 kernel requires $K$ and $N$ to be multiples of 16, and the MXFP8 kernel requires $K$ to be a multiple of 32 and $N$ a multiple of 8.
This excludes 20 call sites of Model A (single-output task heads and one task-head layer with $K\approx3.3\times10^2$) and 3 single-output task heads of Model B.
Unless noted otherwise, we compare all formats on the call sites that every format supports;
MXFP6 supports all call sites.

\textbf{Accuracy metric.}
For each linear call site, we measure the QSNR of Section~\ref{subsec:mxfp8_fidelity} with the open-source QuantSuite framework~\cite{quantsuite2026},
where $y$ is the output of the layer in the evaluated format, $y_{\mathrm{ref}}$ is the FP16 output for the same input,
and the norms run over all captured batches.
We compute $y$ with the actual low-precision GEMM kernels rather than with simulated quantization.
Simulated (fake-quantized) MXFP8 and MXFP6 references agree with the kernels to within 0.15~dB at the median.
As a reference point, we also report the QSNR of FP16 against an FP32 computation, which measures the rounding noise of the FP16 baseline itself.
MXFP6 QSNR is from our production MXFP6 kernel; the deep-pipelined kernel whose performance we report (Section~\ref{subsec:mxfp6_perf}) is 0.04--0.43~dB more accurate on the largest shapes, so the reported QSNR is slightly conservative.

\textbf{Performance metric.}
We measure the GEMM kernel time per call on an AMD MI350X GPU, for batch sizes of 1024 and 2048.
We measured all formats on the same day with the same software build, each at the real $M$ of every layer.
The FP16 baseline uses PyTorch's \texttt{F.linear}.
The FP8, MXFP8, and MXFP6 measurements time the low-precision GEMM kernel on pre-quantized activations,
so they exclude the computational cost of quantizing activations, which we discuss in Sections~\ref{subsec:mxfp8_perf} and~\ref{subsec:mxfp6_perf}.
We flush the L2 cache before each timed call and report the median of 50 calls.
For each shape, we use the fastest configuration of the MXFP8 kernel.
For MXFP6, we report our deep-pipelined kernel (Section~\ref{subsec:mxfp6_perf}) with activations stored K-tile-major;
for each shape, we search all of its configurations combined with split-K factors from 1 to 16,
keeping only combinations whose output matches an FP32 reference.
The GEMM time of the whole dense network is the sum of the per-call times over all call sites.

\subsection{Accuracy and Performance Across All Linear Layers}
\label{sec:eval:all}

\begin{table*}[t]
\caption{QSNR distribution and GEMM speedup over FP16 for the 20 most compute-intensive linear layers and for all linear layers of the two production models.
For accuracy only, the FP16 row reports QSNR against FP32; the low-precision rows report QSNR against FP16.
All speedups use FP16 execution time as the baseline, so FP16 is $1.00\times$ by definition.
MXFP6 speedups are those of the deep-pipelined kernel.
Total speedup is the ratio of the summed GEMM time; per-shape speedup is the geometric mean over unique shapes.
The all-layer blocks cover the call sites that every format supports.}
\label{tab:eval_all}
\centering
\small
\begin{tabular}{@{}lcccccccccc@{}}
\toprule
 & \multicolumn{4}{c}{QSNR (dB)} & \multicolumn{3}{c}{Speedup over FP16} & \multicolumn{3}{c}{Speedup over FP16} \\
 & \multicolumn{4}{c}{} & \multicolumn{3}{c}{Batch 1024} & \multicolumn{3}{c}{Batch 2048} \\
\cmidrule(lr){2-5} \cmidrule(lr){6-8} \cmidrule(l){9-11}
Format & Min & P10 & Median & Mean & Total & Per-shape & Shapes $<1\times$ & Total & Per-shape & Shapes $<1\times$ \\
\midrule
\multicolumn{11}{@{}l}{\emph{Model A, 20 most compute-intensive linear layers (6 shapes)}} \\
FP16 & 70.9 & 71.1 & 71.7 & 71.8 & $1.00\times$ & $1.00\times$ & -- & $1.00\times$ & $1.00\times$ & -- \\
FP8 (row)        & 32.5 & 33.1 & 34.6 & 35.5 & $1.64\times$ & $1.68\times$ & 0 & $1.82\times$ & $1.73\times$ & 0 \\
MXFP8            & 32.3 & 32.8 & 34.6 & 35.4 & $1.67\times$ & $1.71\times$ & 0 & $1.70\times$ & $1.68\times$ & 0 \\
MXFP6            & 30.1 & 30.9 & 32.9 & 33.7 & $2.27\times$ & $2.19\times$ & 0 & $2.29\times$ & $2.17\times$ & 0 \\
\midrule
\multicolumn{11}{@{}l}{\emph{Model A, all linear layers (293 call sites, 45 shapes)}} \\
FP16 & 64.5 & 70.5 & 71.9 & 71.8 & $1.00\times$ & $1.00\times$ & -- & $1.00\times$ & $1.00\times$ & -- \\
FP8 (row)        & 23.2 & 32.0 & 35.6 & 36.1 & $1.49\times$ & $1.29\times$ & 3 & $1.63\times$ & $1.34\times$ & 0 \\
MXFP8            & 23.6 & 31.9 & 35.4 & 35.9 & $1.43\times$ & $1.23\times$ & 3 & $1.55\times$ & $1.31\times$ & 2 \\
MXFP6            & 22.3 & 30.1 & 34.0 & 34.8 & $1.83\times$ & $1.38\times$ & 1 & $1.98\times$ & $1.46\times$ & 0 \\
\midrule
\multicolumn{11}{@{}l}{\emph{Model B, 20 most compute-intensive linear layers (9 shapes)}} \\
FP16 & 73.6 & 73.7 & 73.7 & 73.7 & $1.00\times$ & $1.00\times$ & -- & $1.00\times$ & $1.00\times$ & -- \\
FP8 (row)        & 28.6 & 31.8 & 38.5 & 38.8 & $1.58\times$ & $1.57\times$ & 0 & $1.72\times$ & $1.59\times$ & 0 \\
MXFP8            & 28.2 & 31.7 & 38.3 & 38.8 & $1.72\times$ & $1.61\times$ & 0 & $1.69\times$ & $1.61\times$ & 0 \\
MXFP6            & 26.7 & 30.5 & 36.8 & 37.7 & $2.21\times$ & $2.00\times$ & 0 & $2.31\times$ & $2.13\times$ & 0 \\
\midrule
\multicolumn{11}{@{}l}{\emph{Model B, all linear layers (199 call sites, 45 shapes)}} \\
FP16 & 72.6 & 73.6 & 73.7 & 73.7 & $1.00\times$ & $1.00\times$ & -- & $1.00\times$ & $1.00\times$ & -- \\
FP8 (row)        & 28.5 & 29.0 & 36.6 & 36.4 & $1.49\times$ & $1.30\times$ & 8 & $1.63\times$ & $1.35\times$ & 4 \\
MXFP8            & 28.0 & 28.7 & 36.4 & 36.3 & $1.33\times$ & $1.04\times$ & 18 & $1.49\times$ & $1.12\times$ & 15 \\
MXFP6            & 26.7 & 28.1 & 35.0 & 35.3 & $1.73\times$ & $1.35\times$ & 4 & $1.99\times$ & $1.46\times$ & 2 \\
\bottomrule
\end{tabular}
\end{table*}

Table~\ref{tab:eval_all} summarizes the accuracy and performance of each format across all linear layers and across the 20 most compute-intensive layers (Section~\ref{sec:eval:top20}).
FP16 itself is 72~dB (Model A) to 74~dB (Model B) away from an FP32 computation at the median,
so the noise that the low-precision formats add dominates the noise of the baseline by more than 35~dB.

The accuracy results are consistent across the two models.
Row-wise FP8 and MXFP8 have nearly the same accuracy:
their median QSNRs are 35.6~dB and 35.4~dB for Model A and 36.6~dB and 36.4~dB for Model B,
and the median per-call-site gap between them is only 0.1~dB.
MXFP6 has lower median QSNRs of 34.0~dB and 35.0~dB.
The median per-call-site gap between MXFP6 and MXFP8 is 1.0~dB (Model A) and 0.9~dB (Model B);
for Model A, it is between 0.2~dB and 3.1~dB below MXFP8 at every call site,
and for Model B, it ranges from 3.9~dB below to 1.6~dB above MXFP8.
This gap is far smaller than the 6~dB per mantissa bit of Section~\ref{subsec:mxfp8_fidelity},
which is consistent with E2M3 keeping all three mantissa bits of E4M3:
MXFP6 gives up exponent range, not precision.
The lowest QSNRs appear in small layers of the user-sequence modules (22.3--23.6~dB in Model A, 28.0--28.5~dB in Model B),
except for MXFP6 in Model B, whose minimum (26.7~dB) is in the bottom MLP.

All three low-precision formats reduce the total GEMM time of both dense networks, and MXFP6 reduces it the most.
For Model A, FP8, MXFP8, and MXFP6 are $1.49\times$, $1.43\times$, and $1.83\times$ faster than FP16 at batch 1024,
and $1.63\times$, $1.55\times$, and $1.98\times$ faster at batch 2048.
For Model B, they are $1.49\times$, $1.33\times$, and $1.73\times$ faster at batch 1024, and $1.63\times$, $1.49\times$, and $1.99\times$ faster at batch 2048.
MXFP6 is thus $1.22$--$1.23\times$ faster than the production FP8 scheme for Model A and $1.17$--$1.22\times$ faster for Model B.
The per-shape speedups are much lower than the totals:
many small shapes gain little from low precision, while the large shapes that dominate the GEMM time gain the most.
For Model B, MXFP8 is even slower than FP16 on 18 of the 45 shapes at batch 1024, while MXFP6 is slower on only 4.

Appendix~\ref{sec:appendix:gemm_size} provides the speedup breakdown by relative GEMM size and discusses the performance of request-level layers.

\subsection{Detailed Results for the Largest Linear Layers}
\label{sec:eval:top20}

\begin{table}[t]
\caption{Linear shapes of the 20 most compute-intensive layers of each model, grouped by component. $K$ is the input dimension and $N$ the output dimension; ranges indicate multiple shapes. Dimensions are rounded to two significant digits; $a\mathrm{e}b$ denotes $a\times10^b$.}
\label{tab:eval_top20}
\centering
\footnotesize
\setlength{\tabcolsep}{2pt}
\begin{tabular}{@{}llrcc@{}}
\toprule
Component & Label & Layers & $K$ & $N$ \\
\midrule
\multicolumn{5}{@{}l}{\emph{Model A}} \\
MLP, wide input & MLP-W & 1 & $1.6\mathrm{e}5$ & $6.1\mathrm{e}3$ \\
MLP & MLP & 7 & $2.3\mathrm{e}4$ & $6.1\mathrm{e}3$ \\
Projection & Proj. & 9 & $6.1\mathrm{e}3$ & $3.3\mathrm{e}4$ \\
DCN down-proj. & DCN-D & 1 & $1.6\mathrm{e}5$ & $1.0\mathrm{e}3$ \\
DCN up-proj. & DCN-U & 1 & $1.0\mathrm{e}3$ & $1.6\mathrm{e}5$ \\
DCPP & DCPP & 1 & $1.0\mathrm{e}3$ & $1.5\mathrm{e}5$ \\
\midrule
\multicolumn{5}{@{}l}{\emph{Model B}} \\
MLP & MLP & 6 & $1.5\mathrm{e}4\text{--}6.8\mathrm{e}4$ & $3.1\mathrm{e}3\text{--}6.1\mathrm{e}3$ \\
Projection & Proj. & 5 & $6.1\mathrm{e}3$ & $2.5\mathrm{e}4$ \\
Low-rank down & LR-D & 2 & $2.5\mathrm{e}4$ & $1.0\mathrm{e}3$ \\
Low-rank up & LR-U & 5 & $1.0\mathrm{e}3$ & $3.7\mathrm{e}4\text{--}6.5\mathrm{e}4$ \\
Bottom MLP & Bot. & 2 & $3.1\mathrm{e}3\text{--}1.2\mathrm{e}4$ & $3.1\mathrm{e}3\text{--}8.2\mathrm{e}3$ \\
\bottomrule
\end{tabular}
\end{table}

We call the 20 linear layers with the most GEMM FLOPs per inference the 20 most compute-intensive layers of a model.
They account for 67\% (Model A) and 76\% (Model B) of the GEMM FLOPs
and for 49--54\% (Model A) and 55--59\% (Model B) of the FP16 GEMM time,
so most of the overall GEMM speedup comes from them.
For Model A, they are also the 20 largest linear layers by parameter count and hold 68\% of the parameters of all linear layers;
for Model B, they hold 49\% of the parameters.
Except for two layers of Model B's bottom MLP, all of them belong to the feature-interaction backbone (Section~\ref{sec:background:arch}).
Table~\ref{tab:eval_top20} groups them by component:
the first linear of the MLP in each interaction layer (\emph{MLP}; in Model A, the much wider MLP of the first layer, which reads the full concatenated input, is \emph{MLP-W}),
the projection of each interaction layer's output back to embeddings (\emph{Proj.}),
the low-rank down- and up-projections of the DCN in Model A (\emph{DCN-D} and \emph{DCN-U}) and inside the interaction layers of Model B (\emph{LR-D} and \emph{LR-U}),
the output layer of Model A's DCPP weight network (\emph{DCPP}),
and the first two layers of Model B's bottom MLP, which processes the dense features (\emph{Bot.}).

Compared to all layers, the most compute-intensive layers of Model A have a narrower and lower QSNR distribution (Table~\ref{tab:eval_all}):
the small, sensitive layers that set the minimum of the whole model are absent,
but the median QSNR is lower for every format (34.6~dB for FP8 and MXFP8, and 32.9~dB for MXFP6).
For Model B, the median QSNR of the most compute-intensive layers is higher than that of all layers (38.5~dB for FP8 and 36.8~dB for MXFP6).
The QSNR loss of MXFP6 is also somewhat larger on these layers:
its median per-call-site gap to MXFP8 is 1.8~dB for Model A and 1.2~dB for Model B, compared to 1.0~dB and 0.9~dB across all layers.
In exchange, these layers gain much more from low precision.
No format is slower than FP16 on any of their shapes, and MXFP6 is the fastest format for every one of them in both models and at both batch sizes.
MXFP6 reduces their total GEMM time by $2.27\times$ and $2.29\times$ for Model A and by $2.21\times$ and $2.31\times$ for Model B at batch 1024 and 2048,
compared to $1.58$--$1.82\times$ for FP8 and $1.67$--$1.72\times$ for MXFP8.

The accuracy ordering of the whole model holds for each of these layers (Appendix~\ref{sec:appendix:top20} plots the QSNR of every layer):
FP8 and MXFP8 are within 0.5~dB (Model A) and 0.9~dB (Model B) of each other,
and MXFP6 is 1.0~dB to 2.4~dB (Model A) and 0.6~dB to 1.8~dB (Model B) below MXFP8.
The most sensitive components differ between the models:
the MLP and projection layers in Model A (MXFP6 QSNRs of 30.1~dB to 35.0~dB),
and the low-rank up-projections and the bottom MLP in Model B (26.7~dB to 36.8~dB).
Across components, MXFP6 gains the most on layers with wide inputs, reaching $2.85\times$ at batch 1024 and $2.90\times$ at batch 2048 for DCN-D in Model A.
It gains the least on the layers with $K\approx1.0\times10^3$ and very wide outputs (DCN-U, DCPP, and LR-U, $1.59$--$1.74\times$ across the two batch sizes).
Table~\ref{tab:eval_top20_perf} and Figure~\ref{fig:eval_top20} in Appendix~\ref{sec:appendix:top20} give the speedups of every component.
For the most compute-intensive layers, MXFP6 therefore trades a median of 1.2--1.8~dB of QSNR per call site relative to FP8 and MXFP8 for 26--40\% more GEMM throughput than FP8.

\subsection{End-to-End Performance}
\label{sec:eval:e2e}

\begin{table*}[t]
\caption{Summary of MXFP6 accuracy and end-to-end performance.
Percentages use all unique linear-layer weights as the denominator, excluding biases and embeddings.
QPS improvements are over the production FP8 configurations at batches 1024 / 2048, respectively.}
\label{tab:eval_e2e}
\centering
\small
\setlength{\tabcolsep}{3pt}
\begin{tabular}{@{}lp{0.25\textwidth}p{0.23\textwidth}p{0.28\textwidth}@{}}
\toprule
Model & Parameters quantized to MXFP6 & E2E QPS improvement & Tasks within 0.01\% NE tolerance \\
\midrule
Model A & 80.8\% & +5.3\% / +1.5\% & 6 / 6 \\
Model B & 72.9\% & +14.3\% / +7.6\% & 3 / 3 \\
\bottomrule
\end{tabular}
\end{table*}

\textbf{Model quality and replacement coverage.}
Our accuracy experiments validate MXFP6 replacement rules covering 80.8\% of Model A's and 72.9\% of Model B's linear-layer weights (Table~\ref{tab:eval_e2e}),
which are also the selections used for end-to-end performance.
The observed relative NE increase remains within the tight 0.01\% tolerance for every evaluated prediction task: six in Model A and three in Model B.
We calculate these percentages directly from the linear inventories and replacement rules, counting each unique weight matrix once.

\textbf{End-to-end throughput.}
Table~\ref{tab:eval_e2e} reports throughput from E2E GPU model performance measurements on MI350X with AOTInductor.
For each model, we compare its production FP8 configuration with a selective configuration that executes 27 (Model A) or 49 (Model B) GEMMs per forward in MXFP6,
with activation preparation fused into the producers by Inductor and FlyDSL MXFP6 GEMMs with batch-specific configurations.
Both configurations of a comparison run on identical request-batched inputs with 1024 or 2048 ads.
Each timing covers the full model, including activation preparation, GEMMs, reductions, and host/launch overhead,
and the reported QPS gain is the geometric mean of seven paired FP8/MXFP6 latency ratios minus one.
Both variants pass the standard benchmark numerical accuracy check at relative tolerance 0.01 against the unchanged original model
(FP32 for Model A, FP16 for Model B); this output-agreement check is separate from an NE evaluation.
Appendix~\ref{sec:appendix:e2e} gives the full benchmark configuration.

MXFP6 improves full-model throughput at both batch sizes, and every timing pair favors it.
The gains decrease with batch size: $5.3\%$ to $1.5\%$ for Model A and $14.3\%$ to $7.6\%$ for Model B.

\textbf{Closing the kernel-to-model performance gap.}
The end-to-end experiments use row-major packed activations, while our improved MXFP6 GEMM achieves its largest gains with K-tile-major activations.
Producer and quantization fusions must generate this layout directly to realize those gains without a separate layout conversion.
Until these fusions are integrated into the production workflow, the per-linear kernel speedups do not translate fully into model throughput.
The reported QPS improvements therefore demonstrate gains achievable with the current integration.
The activation preparation work described in Appendix~\ref{sec:future_work} is expected to improve end-to-end throughput, especially at batch size 2048.
Deploying these K-tile-major producers and preparation fusions across many models requires PyTorch Inductor support, which we are actively adding.

\section{Related Work}
\label{sec:related}

\textbf{Microscaling formats.} The OCP MX specification~\cite{ocp2023mx, rouhani2023mx}, which grew out of the shared-microexponent formats of BDR~\cite{rouhani2023bdr}, attaches one E8M0 scale to every 32 consecutive elements of MXFP8, MXFP6, and MXFP4 tensors. NVIDIA's NVFP4~\cite{nvidia2025nvfp4} halves the block to 16 elements and stores its block scales in E4M3, with a second, per-tensor scale in FP32. For 4-bit LLM inference in these formats, MR-GPTQ~\cite{egiazarian2026mrgptq} applies Hadamard rotations within each block before GPTQ-style rounding and reports W4A4 speedups on NVIDIA Blackwell GPUs. None of these works measures native MXFP6 GEMMs. On Blackwell, MXFP6 has the same peak throughput as FP8 (Table~\ref{table:hw_throughput}); AMD MI350X is, to the best of our knowledge, the first GPU on which MXFP6 has twice the peak throughput of FP8.

\textbf{Quantization of LLMs.} Weight-only LLM quantization methods reduce the memory traffic of reading weights during token generation. GPTQ~\cite{frantar2023gptq}, AWQ~\cite{lin2024awq}, and SignRound~\cite{cheng2024signround} differ in how they choose the low-bit weights, and FP6-LLM~\cite{xia2024fp6llm} carries the approach over to 6-bit floating-point weights. FP6-LLM converts its 6-bit weights to FP16 before matrix multiplication, so its advantage diminishes as GEMMs become compute bound, as the large GEMMs of R\&R inference are. Lowering the precision of the activations as well requires dealing with outlier channels: LLM.int8()~\cite{dettmers2022llmint8} computes them in 16 bits, SmoothQuant~\cite{xiao2023smoothquant} rescales channels so that the weights absorb them, and QuaRot~\cite{ashkboos2024quarot} rotates them away to reach 4 bits. These methods are judged against LLM accuracy tolerances, such as retaining 99\% of the zero-shot accuracy~\cite{ashkboos2024quarot}, which are far looser than the $0.01\%$ NE bar of R\&R models.

\textbf{Quantization for R\&R models.} Quantization of DLRM-style models~\cite{naumov2019dlrm} has mostly targeted the memory-bound embedding tables, through INT8 and INT4 embedding quantization~\cite{park2018deep}, tensor-train decomposition~\cite{yin2021ttrec}, and adaptive embedding dimensions~\cite{zhao2021autodim}; these techniques reduce the memory footprint but not the compute of the dense layers. Early INT8 quantization of MLPs~\cite{deng2021lowprecision} often led to unacceptable click-through-rate (CTR) degradation under the strict $0.01\%$ NE launch bar of ads models. More recently, TokenMixer-Large~\cite{tokenmixer2026} and OneRec-V2~\cite{onerec2026} applied FP8~\cite{micikevicius2022fp8} selectively, to Mixture-of-Experts feed-forward layers and generative recommendation models, respectively. LoKA~\cite{luo2026loka} showed that synthetic benchmarks underestimate the quantization error of production models and systematized selective FP8 with distribution-aware profiling, architectural adaptations, and per-operator fallback to FP16. All of these works use per-tensor or per-row scaling and stay at 8 bits or above. Block-wise MX scaling helps most when the values within a row span a wide dynamic range; on our production models, row-wise FP8 and MXFP8 have nearly the same accuracy (Section~\ref{sec:eval:all}), so MXFP8 is a viable alternative to row-wise FP8, chosen by kernel performance.

\section{Conclusion}
\label{sec:conclusion}

In this paper, we addressed the challenge of scaling inference throughput for accuracy-sensitive Recommendation and Ranking (R\&R) models while strictly maintaining model quality within a $0.01\%$ Normalized Entropy (NE) margin relative to the production baselines. We showed that MXFP8 is a viable alternative to row-wise FP8 for industrial R\&R models, with nearly the same accuracy, and pushed sub-8-bit quantization further through selective MXFP6 deployment on the AMD MI350X architecture. From an efficiency perspective, our implementation achieves measured GEMM speedups over FP16 of $1.33$--$1.55\times$ for MXFP8 and $1.73$--$1.99\times$ for MXFP6 across all supported linear layers, and up to $2.90\times$ for MXFP6 on individual layers. Full-model benchmarks at batches 1024 and 2048 show end-to-end throughput gains over the FP8 configurations of $5.3\%$ and $1.5\%$ for Model A and $14.3\%$ and $7.6\%$ for Model B, respectively. To the best of our knowledge, this is the first work to deploy native MXFP6 GEMMs on R\&R models, delivering significant speedups within the $0.01\%$ NE launch bar.

\bibliography{example_paper}
\bibliographystyle{mlsys2025}

\section*{AI Usage Disclosure}
AI tools contributed to this work in two ways. First, as part of the kernel development described in Sections~\ref{subsec:mxfp8_perf} and~\ref{subsec:mxfp6_perf}, the AI agent AlphaKernel~\cite{meta_kernelevolve_2026} tuned the parameters of the GEMM kernel configurations that the authors designed and implemented. Second, the authors used AI assistants (Muse Code, OpenAI Codex, Claude Code, and Google Gemini) to draft, edit, and condense the manuscript, survey and verify related work, cross-check the numbers reported in the text against the measured data, write the scripts that process benchmark results and generate tables and figures, and launch and track benchmark runs. The authors designed the kernels and the experiments, reviewed every AI-assisted change to the text and code, and take full responsibility for the content and results of this paper.

\appendix

\section{Appendix}
\label{sec:appendix}

\subsection{QSNR Gain per Mantissa Bit}
\label{sec:appendix:qsnr_bit}

Section~\ref{subsec:mxfp8_fidelity} states that each additional mantissa bit increases QSNR by about 6~dB. Under standard quantization noise theory~\cite{oppenheim2009discrete, gray1998quantization}, assuming a uniform distribution of values within a quantization bin of width $\Delta$, the quantization noise variance is given by:
\begin{equation}
\sigma_e^2 = \frac{\Delta^2}{12}
\end{equation}
For a floating-point interval $[2^E, 2^{E+1})$, adding $1$ bit of mantissa reduces the bin width from $\Delta = 2^{E-M}$ to $\Delta' = 2^{E-(M+1)} = \frac{\Delta}{2}$. The resulting noise variance becomes:
\begin{equation}
\sigma_{e'}^2 = \frac{(\Delta/2)^2}{12} = \frac{1}{4} \sigma_e^2
\end{equation}
Expressing this error reduction in decibels ($\text{dB}$) demonstrates that each additional mantissa bit increases QSNR by approximately $6.02\text{ dB}$:
\begin{equation}
\Delta \text{QSNR} = 10 \log_{10} \left( \frac{\sigma_e^2}{\sigma_{e'}^2} \right) = 10 \log_{10}(4) \approx 6.02 \text{ dB}
\end{equation}

\subsection{Overflow-Aware Scaling}
\label{sec:appendix:oas}

This section details the Overflow-Aware Scaling (OAS) algorithm of~\citet{mxfp4_oas} and its adaptations to MXFP8 (Section~\ref{subsec:mxfp8_fidelity}) and MXFP6 (Section~\ref{subsec:mxfp6_fidelity}). For MXFP8, the standard scale computation maps the per-block absolute maximum $\alpha_{\max}$ to the representable range $(224, 448]$, which prevents saturation (clamping) error. OCP microscaling (MXFP8-OCP) instead maps $\alpha_{\max}$ to $(256, 512]$, which allows some overflow above $F_{\max} = 448$. OAS maps $\alpha_{\max}$ to a range whose upper bound is the midpoint between $F_{\max}$ and the next value of the element grid. Compared with the standard computation, OAS accepts a small saturation of the block maximum in exchange for doubling the representable dynamic range for the lower-magnitude elements of the affected blocks. Compared with MXFP8-OCP, it avoids severe clamping error at the top of the range~\cite{mxfp4_oas}.

Applied naively to E4M3, OAS ignores the NaN encoding: it treats 480 as the largest value and 512 as the next grid value, and maps $\alpha_{\max}$ to $(248, 496]$. However, the all-ones pattern ($E=1111, M=111$), which would encode 480, is reserved for NaN in E4M3. As a result, values in $(464, 496]$ that would round to 480 must instead be clamped to $F_{\max} = 448$, incurring significant clamping error. We therefore use Overflow-Aware Scaling Skip (OAS-S), an adaptation of OAS~\cite{mxfp4_oas} to E4M3 that skips the all-ones pattern during scale rounding: it applies OAS with the true $F_{\max} = 448$ and uses 464, the midpoint between 448 and 480, as the upper bound. OAS-S maps $\alpha_{\max}$ to $(232, 464]$, so mapped values never round to the NaN encoding, and the relative error from saturating the block maximum is at most $(464 - 448)/464 = 3.4\%$.

For example, if $\alpha_{\max}$ maps to 472 under MXFP8-OCP, both MXFP8-OCP and naive OAS keep the E8M0 scale exponent at 0 (a scale of 1). At the quantization step, this value is clamped to 448, yielding a relative quantization error of $\left|\frac{472 - 448}{472}\right| = 5.08\%$. Under OAS-S, scaling 472 down by a factor of 2 maps it to 236, which rounds to the nearest representable E4M3 value of 240 with a relative error of only $\left|\frac{236 - 240}{236}\right| = 1.69\%$. OAS-S thus avoids the NaN encoding while substantially reducing quantization error.

For MXFP6, E2M3 has no NaN encoding, so we apply the regular OAS algorithm. OAS maps $\alpha_{\max}$ to $(3.875, 7.75]$, compared with $(3.75, 7.5]$ for the standard computation and $(4, 8]$ for MXFP6-OCP. For example, if $\alpha_{\max}$ maps to $7.9$ under MXFP6-OCP, it is clamped to $F_{\max} = 7.5$, a relative error of $\left|\frac{7.9 - 7.5}{7.9}\right| = 5.06\%$. OAS instead maps it to $3.95$, which rounds to $4.0$ with a relative error of only $\left|\frac{3.95 - 4.0}{3.95}\right| = 1.27\%$.

\subsection{GEMM Speedup by Relative Size}
\label{sec:appendix:gemm_size}

This section expands the all-layer performance analysis in Section~\ref{sec:eval:all}.

\begin{table*}[t]
\caption{GEMM speedup by relative GEMM size, for the call sites that every format supports.
FP16 time is the share of the total FP16 GEMM time; the FP8, MXFP8, and MXFP6 columns give the speedup over FP16, and the last column of each batch size gives the speedup of MXFP6 over MXFP8.
Work per call is normalized to the largest supported shape of each model at batch 1024.
The four bands have boundaries at approximately 0.05\%, 0.50\%, and 4.98\% of that reference for Model A,
and 0.12\%, 1.18\%, and 11.76\% for Model B. Shapes keep the same band at batch 2048.}
\label{tab:eval_size}
\centering
\small
\setlength{\tabcolsep}{4pt}
\begin{tabular}{@{}lrcccccccccc@{}}
\toprule
 & & \multicolumn{5}{c}{Batch 1024} & \multicolumn{5}{c}{Batch 2048} \\
\cmidrule(lr){3-7} \cmidrule(l){8-12}
Work band & Shapes & \begin{tabular}[b]{@{}c@{}}FP16\\ time\end{tabular} & FP8 & MXFP8 & MXFP6 & \begin{tabular}[b]{@{}c@{}}MXFP6 vs.\\ MXFP8\end{tabular} & \begin{tabular}[b]{@{}c@{}}FP16\\ time\end{tabular} & FP8 & MXFP8 & MXFP6 & \begin{tabular}[b]{@{}c@{}}MXFP6 vs.\\ MXFP8\end{tabular} \\
\midrule
\multicolumn{12}{@{}l}{\emph{Model A}} \\
Smallest & 11 & 4.4\%  & 1.05 & 1.13 & 1.17 & 1.03 & 2.9\%  & 1.07 & 1.18 & 1.20 & 1.02 \\
Small    & 16 & 8.0\%  & 1.23 & 1.00 & 1.17 & 1.16 & 6.5\%  & 1.32 & 1.17 & 1.28 & 1.09 \\
Medium   & 12 & 34.1\% & 1.46 & 1.31 & 1.65 & 1.26 & 33.3\% & 1.50 & 1.50 & 1.86 & 1.23 \\
Largest  & 6  & 53.5\% & 1.62 & 1.66 & 2.29 & 1.38 & 57.2\% & 1.81 & 1.67 & 2.29 & 1.37 \\
\midrule
\multicolumn{12}{@{}l}{\emph{Model B}} \\
Smallest & 16 & 11.4\% & 1.09 & 0.79 & 1.10 & 1.40 & 8.0\%  & 1.12 & 0.87 & 1.18 & 1.36 \\
Small    & 13 & 9.6\%  & 1.42 & 0.91 & 1.15 & 1.27 & 6.8\%  & 1.36 & 0.98 & 1.24 & 1.27 \\
Medium   & 12 & 31.7\% & 1.55 & 1.35 & 1.72 & 1.27 & 34.0\% & 1.66 & 1.65 & 2.08 & 1.26 \\
Largest  & 4  & 47.3\% & 1.60 & 1.76 & 2.30 & 1.31 & 51.1\% & 1.77 & 1.69 & 2.37 & 1.40 \\
\bottomrule
\end{tabular}
\end{table*}

Table~\ref{tab:eval_size} groups the shapes by their size.
The speedup of MXFP6 over FP16 grows with GEMM size in both models.
The shapes in the largest compute band account for about half of the FP16 GEMM time,
and on them MXFP6 is $2.29\times$ faster than FP16 for Model A at both batch sizes and $2.30\times$ and $2.37\times$ faster for Model B at batch 1024 and 2048,
compared to at most $1.81\times$ for FP8 and MXFP8.
On these shapes, MXFP6 is $1.31$--$1.40\times$ faster than MXFP8,
a substantial part of the $2\times$ ratio of their peak throughputs on MI350X (Table~\ref{table:hw_throughput}).
Large GEMMs are compute bound and can use the higher peak throughput of the low-precision formats,
whereas the time of small GEMMs is dominated by fixed overheads that low precision does not reduce.
For Model B, whose small layers take a larger share of the time, MXFP8 is slower than FP16 on the GEMMs in the two smallest compute bands ($0.79$--$0.98\times$),
while MXFP6 still gains $1.10$--$1.24\times$ and FP8 $1.09$--$1.42\times$.
Most of these small GEMMs (19 of 29 shapes) belong to layers that run once per request rather than once per candidate ad,
so their $M$ is only 25 rows at batch 1024 and 51 rows at batch 2048.
On these request-level layers, MXFP6 is $1.11$--$1.16\times$ faster than FP16 in total, but FP8 is faster still ($1.31$--$1.32\times$)
and beats MXFP6 on 14 of the 19 shapes at batch 1024 and on 16 at batch 2048.
No format is therefore best for every layer:
the large GEMMs, which dominate the time, favor MXFP6, while the smallest ones favor FP8.

\subsection{Per-Component Results for the 20 Most Compute-Intensive Layers}
\label{sec:appendix:top20}

\begin{figure*}[t]
\centering
\subfigure[Model A: QSNR of each of the 20 most compute-intensive linear layers.]{%
  \begin{minipage}[b]{0.48\textwidth}\definecolor{series1}{HTML}{2A78D6}
\definecolor{series2}{HTML}{EB6834}
\definecolor{series3}{HTML}{1BAF7A}
\begin{tikzpicture}
\begin{axis}[
    width=\linewidth, height=4.6cm,
    ymin=28, ymax=48, ytick={30,35,40,45},
    ylabel={QSNR vs.\ FP16 (dB)}, ylabel style={font=\footnotesize},
    xmin=-0.90, xmax=27.10,
    xtick={0.00,5.60,14.40,21.00,23.60,26.20},
    xticklabels={MLP-W,MLP,Proj.,DCN-D,DCN-U,DCPP},
    xticklabel style={rotate=45, anchor=north east, inner sep=1pt},
    axis x line*=bottom, axis y line*=left,
    ymajorgrids, grid style={line width=0.4pt, draw=black!12},
    tick label style={font=\footnotesize}, tick style={draw=none},
    legend style={font=\footnotesize, draw=none, fill=none, at={(0.5,1.02)}, anchor=south,
                  legend columns=3, column sep=6pt},
    legend cell align=left,
]
\addplot[only marks, mark=*, mark size=2.2pt, mark options={fill=series1, draw=white, line width=0.3pt}]
    coordinates {(-0.34,32.9) (2.26,33.1) (3.26,34.6) (4.26,33.7) (5.26,35.0) (6.26,36.6) (7.26,35.6) (8.26,35.0) (10.06,34.7) (11.06,36.0) (12.06,33.2) (13.06,34.6) (14.06,34.7) (15.06,33.4) (16.06,34.0) (17.06,32.5) (18.06,33.2) (20.66,40.8) (23.26,41.4) (25.86,45.3)};
\addlegendentry{FP8 (row)}
\addplot[only marks, mark=square*, mark size=1.9pt, mark options={fill=series2, draw=white, line width=0.3pt}]
    coordinates {(0.00,32.8) (2.60,32.8) (3.60,34.6) (4.60,33.7) (5.60,35.0) (6.60,36.6) (7.60,35.6) (8.60,34.9) (10.40,34.6) (11.40,35.8) (12.40,33.0) (13.40,34.4) (14.40,34.6) (15.40,33.2) (16.40,33.8) (17.40,32.3) (18.40,33.0) (21.00,40.8) (23.60,41.0) (26.20,45.6)};
\addlegendentry{MXFP8}
\addplot[only marks, mark=triangle*, mark size=2.6pt, mark options={fill=series3, draw=white, line width=0.3pt}]
    coordinates {(0.34,31.2) (2.94,30.4) (3.94,33.5) (4.94,32.0) (5.94,33.7) (6.94,35.0) (7.94,33.9) (8.94,33.1) (10.74,33.3) (11.74,33.7) (12.74,31.2) (13.74,32.7) (14.74,32.8) (15.74,31.1) (16.74,31.8) (17.74,30.1) (18.74,30.9) (21.34,39.8) (23.94,39.7) (26.54,43.6)};
\addlegendentry{MXFP6}
\end{axis}
\end{tikzpicture}\end{minipage}\label{fig:eval_top20_qsnr}}
\hfill
\subfigure[Model A: GEMM speedup over FP16 per component at batch 1024.]{%
  \begin{minipage}[b]{0.48\textwidth}\definecolor{series1}{HTML}{2A78D6}
\definecolor{series2}{HTML}{EB6834}
\definecolor{series3}{HTML}{1BAF7A}
\begin{tikzpicture}
\begin{axis}[
    width=\linewidth, height=4.6cm,
    ybar=1pt, bar width=4.2pt,
    ymin=1, ymax=3, ytick={1,1.5,2,2.5,3},
    yticklabel={$\pgfmathprintnumber{\tick}\times$},
    ylabel={GEMM speedup vs.\ FP16}, ylabel style={font=\footnotesize},
    symbolic x coords={MLP-W,MLP,Proj.,DCN-D,DCN-U,DCPP},
    xtick=data, enlarge x limits=0.1,
    xticklabel style={rotate=45, anchor=north east, inner sep=1pt},
    axis x line*=bottom, axis y line*=left,
    ymajorgrids, grid style={line width=0.4pt, draw=black!12},
    tick label style={font=\footnotesize}, tick style={draw=none},
    legend style={font=\footnotesize, draw=none, fill=none, at={(0.5,1.02)}, anchor=south,
                  legend columns=3, column sep=6pt},
    legend cell align=left,
    legend image code/.code={\draw[#1] (0cm,-0.1cm) rectangle (0.2cm,0.12cm);},
]
\addplot[draw=none, fill=series1] coordinates {(MLP-W,1.8) (MLP,1.4) (Proj.,1.7) (DCN-D,2.5) (DCN-U,1.5) (DCPP,1.4)};
\addlegendentry{FP8 (row)}
\addplot[draw=none, fill=series2] coordinates {(MLP-W,2.0) (MLP,1.6) (Proj.,1.6) (DCN-D,2.1) (DCN-U,1.5) (DCPP,1.6)};
\addlegendentry{MXFP8}
\addplot[draw=none, fill=series3] coordinates {(MLP-W,2.7) (MLP,2.5) (Proj.,2.0) (DCN-D,2.9) (DCN-U,1.6) (DCPP,1.7)};
\addlegendentry{MXFP6}
\end{axis}
\end{tikzpicture}\end{minipage}\label{fig:eval_top20_speedup}}
\\
\subfigure[Model B: QSNR of each of the 20 most compute-intensive linear layers.]{%
  \begin{minipage}[b]{0.48\textwidth}\definecolor{series1}{HTML}{2A78D6}
\definecolor{series2}{HTML}{EB6834}
\definecolor{series3}{HTML}{1BAF7A}
\begin{tikzpicture}
\begin{axis}[
    width=\linewidth, height=4.6cm,
    ymin=25, ymax=51, ytick={30,35,40,45,50},
    ylabel={QSNR vs.\ FP16 (dB)}, ylabel style={font=\footnotesize},
    xmin=-0.90, xmax=25.50,
    xtick={2.50,8.80,13.90,19.00,24.10},
    xticklabels={MLP,Proj.,LR-D,LR-U,Bot.},
    xticklabel style={rotate=45, anchor=north east, inner sep=1pt},
    axis x line*=bottom, axis y line*=left,
    ymajorgrids, grid style={line width=0.4pt, draw=black!12},
    tick label style={font=\footnotesize}, tick style={draw=none},
    legend style={font=\footnotesize, draw=none, fill=none, at={(0.5,1.02)}, anchor=south,
                  legend columns=3, column sep=6pt},
    legend cell align=left,
]
\addplot[only marks, mark=*, mark size=2.2pt, mark options={fill=series1, draw=white, line width=0.3pt}]
    coordinates {(-0.34,46.9) (0.66,48.4) (1.66,40.6) (2.66,46.1) (3.66,40.7) (4.66,42.2) (6.46,38.6) (7.46,45.0) (8.46,36.9) (9.46,42.0) (10.46,34.0) (13.06,34.9) (14.06,41.9) (16.66,31.1) (17.66,34.1) (18.66,38.3) (19.66,37.1) (20.66,36.6) (23.26,28.6) (24.26,31.9)};
\addlegendentry{FP8 (row)}
\addplot[only marks, mark=square*, mark size=1.9pt, mark options={fill=series2, draw=white, line width=0.3pt}]
    coordinates {(0.00,47.7) (1.00,49.2) (2.00,40.6) (3.00,46.6) (4.00,40.8) (5.00,42.4) (6.80,38.6) (7.80,45.2) (8.80,36.8) (9.80,41.7) (10.80,33.6) (13.40,35.0) (14.40,42.3) (17.00,30.5) (18.00,33.7) (19.00,38.0) (20.00,37.1) (21.00,36.5) (23.60,28.2) (24.60,31.8)};
\addlegendentry{MXFP8}
\addplot[only marks, mark=triangle*, mark size=2.6pt, mark options={fill=series3, draw=white, line width=0.3pt}]
    coordinates {(0.34,46.9) (1.34,48.6) (2.34,40.0) (3.34,45.8) (4.34,40.2) (5.34,41.7) (7.14,36.8) (8.14,43.8) (9.14,35.6) (10.14,40.3) (11.14,31.9) (13.74,34.5) (14.74,41.7) (17.34,29.1) (18.34,32.5) (19.34,36.8) (20.34,36.5) (21.34,35.1) (23.94,26.7) (24.94,30.7)};
\addlegendentry{MXFP6}
\end{axis}
\end{tikzpicture}\end{minipage}\label{fig:eval_top20_qsnr_b}}
\hfill
\subfigure[Model B: GEMM speedup over FP16 per component at batch 1024.]{%
  \begin{minipage}[b]{0.48\textwidth}\definecolor{series1}{HTML}{2A78D6}
\definecolor{series2}{HTML}{EB6834}
\definecolor{series3}{HTML}{1BAF7A}
\begin{tikzpicture}
\begin{axis}[
    width=\linewidth, height=4.6cm,
    ybar=1pt, bar width=4.2pt,
    ymin=1, ymax=2.6, ytick={1,1.5,2,2.5},
    yticklabel={$\pgfmathprintnumber{\tick}\times$},
    ylabel={GEMM speedup vs.\ FP16}, ylabel style={font=\footnotesize},
    symbolic x coords={MLP,Proj.,LR-D,LR-U,Bot.},
    xtick=data, enlarge x limits=0.13,
    xticklabel style={rotate=45, anchor=north east, inner sep=1pt},
    axis x line*=bottom, axis y line*=left,
    ymajorgrids, grid style={line width=0.4pt, draw=black!12},
    tick label style={font=\footnotesize}, tick style={draw=none},
    legend style={font=\footnotesize, draw=none, fill=none, at={(0.5,1.02)}, anchor=south,
                  legend columns=3, column sep=6pt},
    legend cell align=left,
    legend image code/.code={\draw[#1] (0cm,-0.1cm) rectangle (0.2cm,0.12cm);},
]
\addplot[draw=none, fill=series1] coordinates {(MLP,1.7) (Proj.,1.4) (LR-D,1.5) (LR-U,1.3) (Bot.,1.7)};
\addlegendentry{FP8 (row)}
\addplot[draw=none, fill=series2] coordinates {(MLP,1.8) (Proj.,1.6) (LR-D,1.3) (LR-U,1.5) (Bot.,1.6)};
\addlegendentry{MXFP8}
\addplot[draw=none, fill=series3] coordinates {(MLP,2.5) (Proj.,2.1) (LR-D,1.6) (LR-U,1.6) (Bot.,2.1)};
\addlegendentry{MXFP6}
\end{axis}
\end{tikzpicture}\end{minipage}\label{fig:eval_top20_speedup_b}}
\caption{Accuracy and performance of the 20 most compute-intensive linear layers of each model, grouped by model component (labels as in Table~\ref{tab:eval_top20}).
Speedup bars start at the FP16 baseline ($1\times$).}
\label{fig:eval_top20}
\end{figure*}
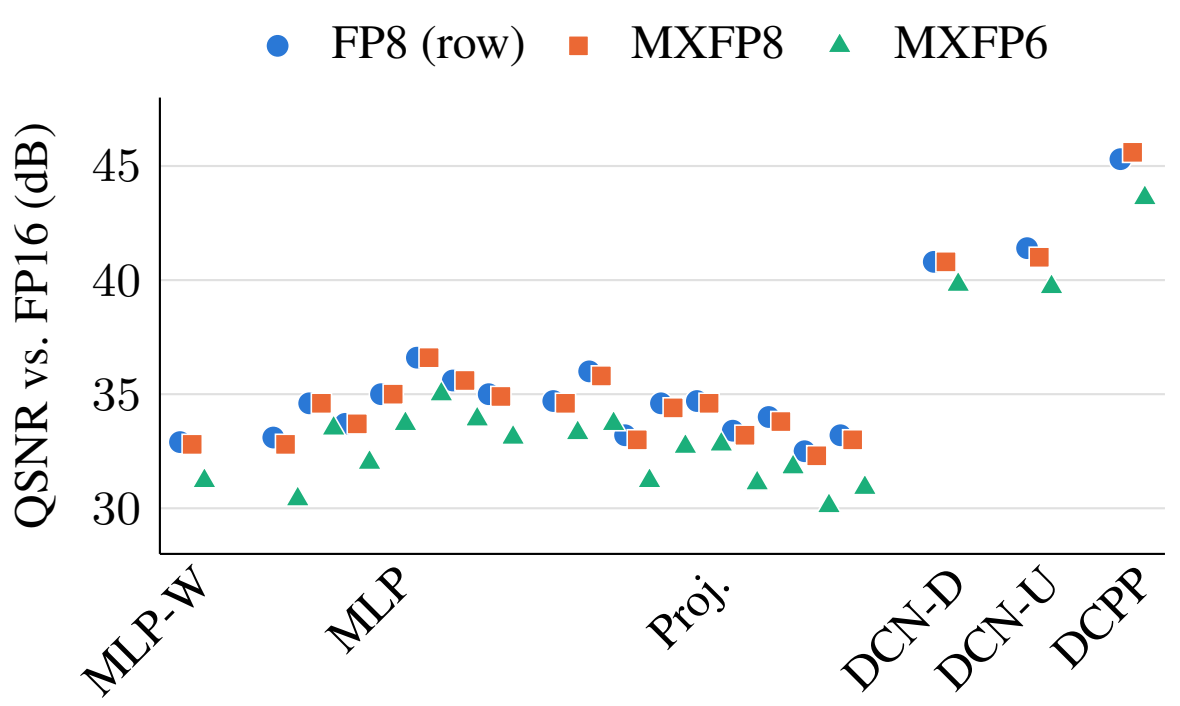
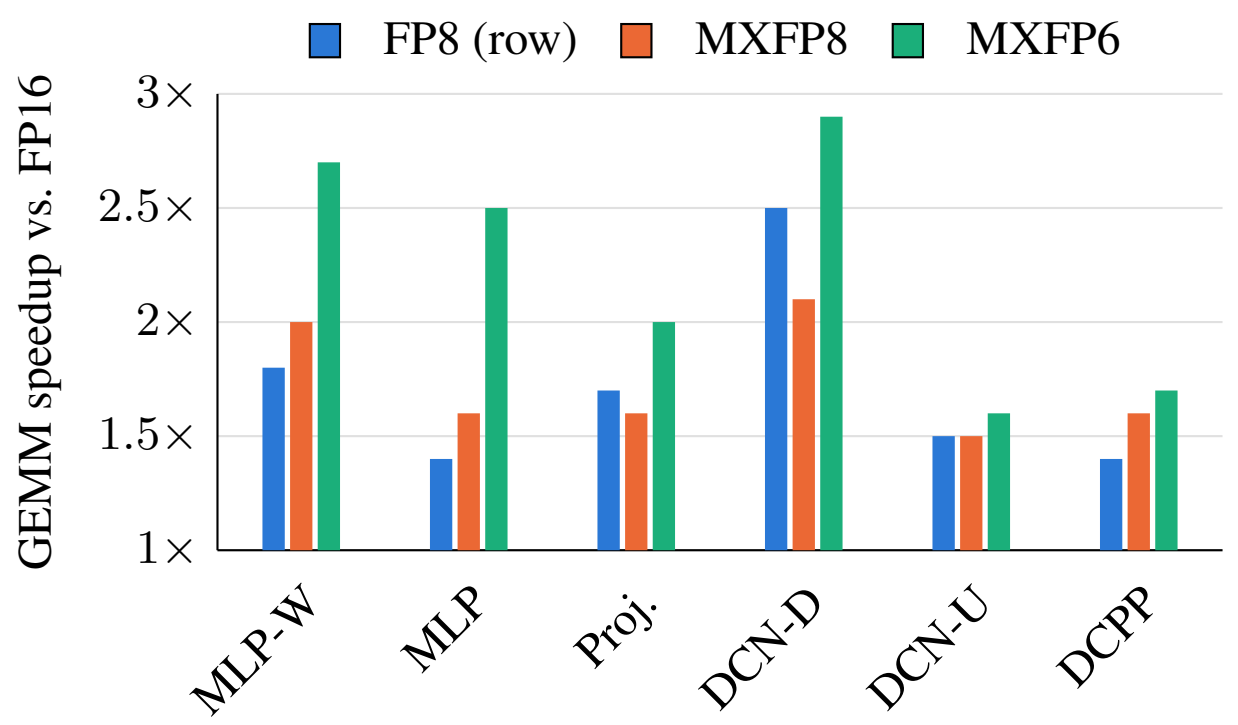
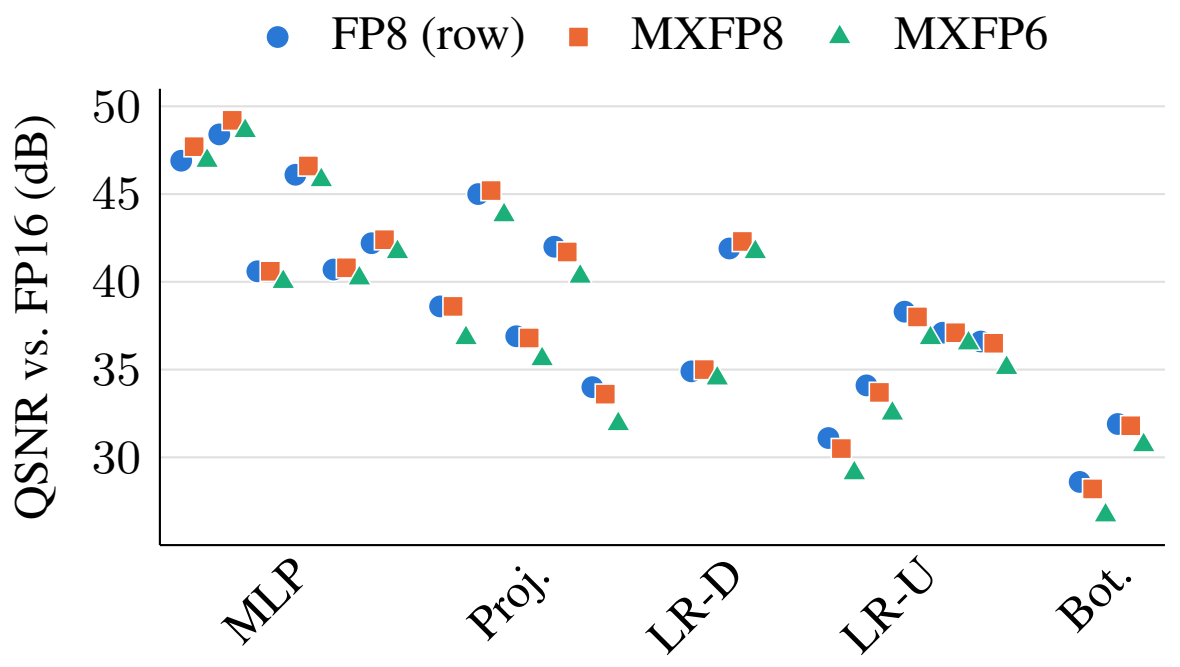
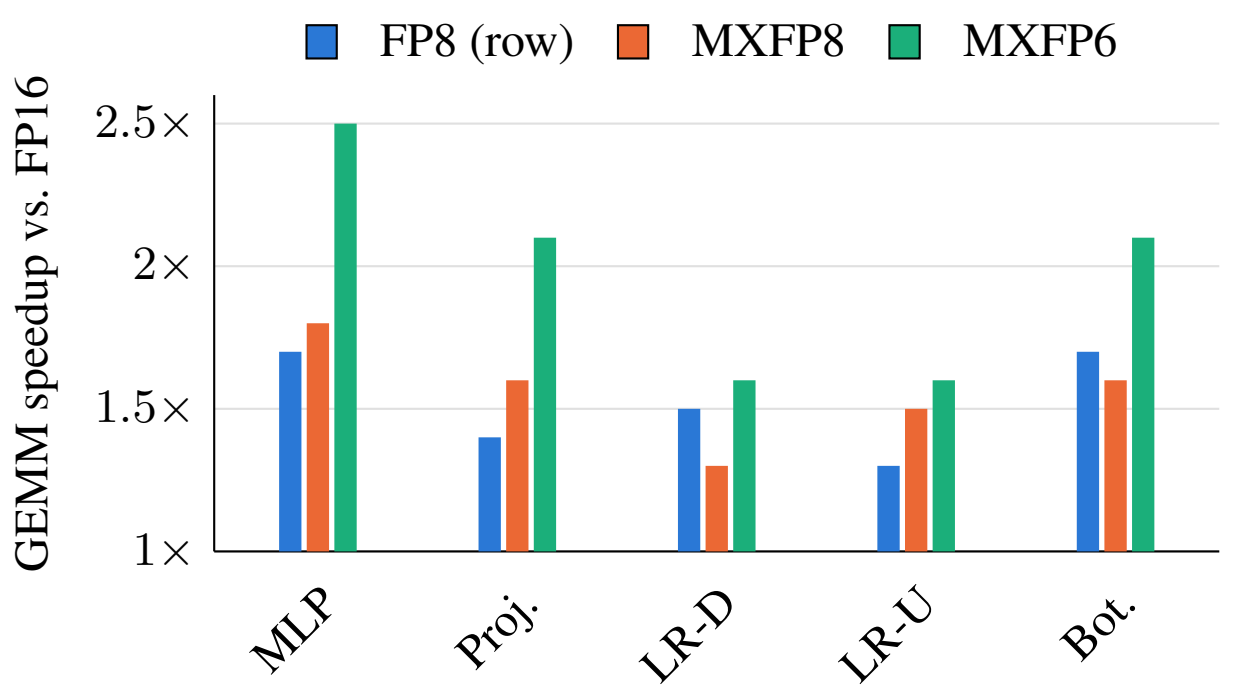

Table~\ref{tab:eval_top20_perf} and Figure~\ref{fig:eval_top20} give the per-component results behind Section~\ref{sec:eval:top20}.
The table reports the mean QSNR of each component and its GEMM speedup over FP16 at both batch sizes.
Panels~(a) and~(c) of the figure show the QSNR of each of the 20 most compute-intensive layers of each model,
and panels~(b) and~(d) show the GEMM speedup over FP16 of each component at batch 1024.

\begin{table*}[t]
\caption{QSNR and GEMM speedup over FP16 for the 20 most compute-intensive linear layers of each model, grouped by model component.
QSNR is the mean over the layers of each component, and speedup is the ratio of their summed GEMM time.
In Model A, all layers of a component share one GEMM shape; in Model B, some components span several shapes.
Linear dimensions are rounded to two significant digits; $a\mathrm{e}b$ denotes $a\times10^b$.}
\label{tab:eval_top20_perf}
\centering
\footnotesize
\setlength{\tabcolsep}{2pt}
\begin{tabular}{@{}llrrrccccccccc@{}}
\toprule
 & & & & & \multicolumn{3}{c}{QSNR (dB)} & \multicolumn{3}{c}{Speedup, batch 1024} & \multicolumn{3}{c}{Speedup, batch 2048} \\
\cmidrule(lr){6-8} \cmidrule(lr){9-11} \cmidrule(l){12-14}
Component & Label & Layers & $K$ & $N$ & FP8 & MXFP8 & MXFP6 & FP8 & MXFP8 & MXFP6 & FP8 & MXFP8 & MXFP6 \\
\midrule
\multicolumn{14}{@{}l}{\emph{Model A}} \\
MLP, wide input    & MLP-W & 1 & $1.6\mathrm{e}5$ & $6.1\mathrm{e}3$ & 32.9 & 32.8 & 31.2 & 1.82 & 1.96 & \textbf{2.72} & 2.19 & 1.94 & \textbf{2.78} \\
MLP                & MLP   & 7 & $2.3\mathrm{e}4$ & $6.1\mathrm{e}3$ & 34.8 & 34.7 & 33.1 & 1.40 & 1.55 & \textbf{2.46} & 1.63 & 1.32 & \textbf{2.27} \\
Projection         & Proj. & 9 & $6.1\mathrm{e}3$ & $3.3\mathrm{e}4$ & 34.0 & 33.9 & 32.0 & 1.70 & 1.58 & \textbf{2.05} & 1.83 & 1.83 & \textbf{2.19} \\
DCN down-proj.     & DCN-D & 1 & $1.6\mathrm{e}5$ & $1.0\mathrm{e}3$ & 40.8 & 40.8 & 39.8 & 2.52 & 2.11 & \textbf{2.85} & 1.98 & 2.06 & \textbf{2.90} \\
DCN up-proj.       & DCN-U & 1 & $1.0\mathrm{e}3$ & $1.6\mathrm{e}5$ & 41.4 & 41.0 & 39.7 & 1.51 & 1.52 & \textbf{1.62} & 1.44 & 1.55 & \textbf{1.65} \\
DCPP               & DCPP  & 1 & $1.0\mathrm{e}3$ & $1.5\mathrm{e}5$ & 45.3 & 45.6 & 43.6 & 1.38 & 1.62 & \textbf{1.72} & 1.45 & 1.50 & \textbf{1.59} \\
\midrule
All 20 layers      &       & 20 &       &        &      &      &      & 1.64 & 1.67 & \textbf{2.27} & 1.82 & 1.70 & \textbf{2.29} \\
\midrule
\multicolumn{14}{@{}l}{\emph{Model B}} \\
MLP                & MLP   & 6 & $1.5\mathrm{e}4\text{--}6.8\mathrm{e}4$ & $3.1\mathrm{e}3\text{--}6.1\mathrm{e}3$ & 44.1 & 44.5 & 43.8 & 1.71 & 1.84 & \textbf{2.48} & 1.83 & 1.71 & \textbf{2.54} \\
Projection         & Proj. & 5 & $6.1\mathrm{e}3$ & $2.5\mathrm{e}4$ & 39.3 & 39.2 & 37.7 & 1.43 & 1.64 & \textbf{2.12} & 1.70 & 1.69 & \textbf{2.20} \\
Low-rank down      & LR-D  & 2 & $2.5\mathrm{e}4$ & $1.0\mathrm{e}3$ & 38.4 & 38.7 & 38.1 & 1.52 & 1.32 & \textbf{1.64} & 2.00 & 1.84 & \textbf{2.43} \\
Low-rank up        & LR-U  & 5 & $1.0\mathrm{e}3$ & $3.7\mathrm{e}4\text{--}6.5\mathrm{e}4$ & 35.4 & 35.2 & 34.0 & 1.35 & 1.50 & \textbf{1.59} & 1.37 & 1.59 & \textbf{1.74} \\
Bottom MLP         & Bot.  & 2 & $3.1\mathrm{e}3\text{--}1.2\mathrm{e}4$ & $3.1\mathrm{e}3\text{--}8.2\mathrm{e}3$ & 30.2 & 30.0 & 28.7 & 1.70 & 1.59 & \textbf{2.08} & 1.40 & 1.45 & \textbf{2.08} \\
\midrule
All 20 layers      &       & 20 &       &              &      &      &      & 1.58 & 1.72 & \textbf{2.21} & 1.72 & 1.69 & \textbf{2.31} \\
\bottomrule
\end{tabular}
\end{table*}

\subsection{End-to-End Benchmark Configuration}
\label{sec:appendix:e2e}

This section gives the full configuration of the end-to-end benchmarks of Section~\ref{sec:eval:e2e}.

\textbf{Benchmark and configurations.}
Table~\ref{tab:eval_e2e} reports timings from the October 28 E2E GPU model performance measurements on MI350X with AOTInductor
(PyTorch \texttt{2.16}, HIP \texttt{7.0.51831}).
Activation preparation uses Inductor producer fusion, preserving both the packed activation layout and the FP16/BF16 input rounding boundary;
the MXFP6 GEMMs use FlyDSL with batch-specific configurations and split-K factors.
The end-to-end experiments use row-major packed activations; Section~\ref{sec:eval:e2e} explains how this layout limits the kernel gains realized in the full model.
Both variants enable \texttt{triton.nested\_reduction} and \texttt{loop\_ordering\_after\_fusion}.
The selective configurations execute 27 MXFP6 GEMMs per forward in Model A and 49 in Model B,
confirmed by traces of 20 complete forwards per variant.

\textbf{Inputs.}
Model A uses benchmark-generated request-batched (IBB) inputs with 128 requests and 1024 ads at batch 1024
and 256 requests and 2048 ads at batch 2048 (seed 20261028).
Model B uses real IBB inputs with 64 requests and 1024 ads, or 128 requests and 2048 ads.
Within each comparison, the inputs have identical values, shapes, dtypes, and strides.

\textbf{Timing and validation.}
For each model and batch size, we retain all seven timing pairs, alternating FP8 and MXFP6 order on the same GPU.
Each side of a pair uses 20 warmups and 200 timed iterations of the full model, including activation preparation,
GEMMs, reductions, and host/launch overhead.
The reported QPS gain is the geometric mean of the seven FP8/MXFP6 latency ratios minus one.
The 95\% intervals use 10,000 paired bootstrap samples (seed 20261028) and describe within-process timing variation;
they do not cover independent compilation or day-to-day variation.
Paired timing uses \texttt{torch.no\_grad} and the original benchmark timer with its default 40~MiB cache flush;
standard benchmark runs use inference mode and also improve in every case.
Both variants pass the standard benchmark numerical accuracy check at relative tolerance 0.01 against the unchanged original model
(FP32 for Model A, FP16 for Model B). This output-agreement check is separate from an NE evaluation.

At batches 1024 / 2048, the paired-bootstrap 95\% intervals for QPS improvement are +4.8--6.1\% / +1.3--1.8\% for Model A and +13.2--16.0\% / +7.1--8.4\% for Model B.
Additional GEMM and preparation tuning improves Model A throughput at batch 2048 by 1.8\% over the previous MXFP6 configuration (95\% interval: +1.6--2.0\%).
With pointwise autotuning and coordinate descent enabled for both FP8 and tuned MXFP6, the improvement over the matched FP8 baseline is 1.2\% (95\% interval: +1.0--1.4\%); the same seven-pair timing and standard benchmark numerical accuracy checks pass with 27 MXFP6 sites.
This separate tuning comparison uses an autotuned FP8 baseline, whereas Table~\ref{tab:eval_e2e} uses the original compiler settings for both variants.

\subsection{Future Work}
\label{sec:future_work}

We are actively implementing producer and preparation fusions that emit K-tile-major packed activations directly, allowing the improved MXFP6 GEMM to consume them without a layout conversion.
We are also actively adding PyTorch Inductor support for this output layout and the corresponding fusions for different producer graphs, enabling broad deployment across production models.
These activation preparation improvements are expected to improve end-to-end throughput, especially at batch size 2048.
We will evaluate the resulting throughput improvements with the same paired timing and accuracy checks.

\end{document}